\documentclass[reqno]{article}
\usepackage{epsfig}
\usepackage{latexsym}
\usepackage{fullpage}
\usepackage{amsthm,amsmath,amssymb}
\usepackage{graphicx} 
\usepackage{subfigure} 

\usepackage{algorithm}
\usepackage{algorithmic}

\newcommand{\argmin}{\text{argmin}}

\begin{document}
\title{TV-SVM: Total Variation Support Vector Machine\\ for Semi-Supervised Data Classification} 

\author{Xavier Bresson\footnote{Department of Computer Science, City University of Hong Kong, Hong Kong.}
         \and Ruiliang Zhang$^\ast$ }

\date{}

\maketitle
\abstract{We introduce semi-supervised data classification algorithms based on total variation (TV), Reproducing Kernel Hilbert Space (RKHS), support vector machine (SVM), Cheeger cut, labeled and unlabeled data points. We design binary and multi-class semi-supervised classification algorithms. We compare the TV-based classification algorithms with the related Laplacian-based algorithms, and show that TV classification perform significantly better when the number of labeled data is small.
}


\section{Introduction}
\subsection{Notation}
Let $\{x_i, y_i\}_{1\leq i \leq N}$ denote $N$ data points, where $x_i\in\mathbb{R}^d$ is its attributes with dimension $d$, while $y_i\in\{+1,-1\}$ (binary classification) or $y_i\in\{1,...,c\}$ (multi-class classification). The total number of data points is $N$ including $n$ labeled data and $N-n$ unlabeled data. $H_K$ is a Reproducing Kernel Hilbert Space (RKHS) with $K:\mathbb{R}^{d\times d}\rightarrow Sym(\mathbb{R})$ is an operator-valued, positive definite kernel. Finally, we use the abbreviation $f_i=f(x_i)$.


\section{Binary (two-class) data classification}


\subsection{Regularized Least Square (RLS)}

The standard RLS problem for binary classification is as follows \cite{art:PoggioSmale03RSVM}. Find a function $f:\mathbb{R}^d\rightarrow\mathbb{R}$ such that
\begin{eqnarray}
\min_{f\in H_K}\ \frac{\eta}{2} \sum_{i\in n} (y_i-f_i)^2 + \frac{\lambda}{2} \|f\|^2_{H_K},
\label{eq:RLS}
\end{eqnarray}
where $\eta,\lambda>0$. Representer theorem states the existence of a minimizing function $f^\star(x)=\sum_{j\in n} K(x,x_j)\alpha_j^\star$ (or $f(x)=K_x\alpha$ with matrix representation) and the norm of $f$ in the RKHS is $\|f\|^2_{H_K}=\alpha^T K \alpha$. Problem \eqref{eq:RLS} is equivalent to
\begin{eqnarray}
\min_{\alpha\in\mathbb{R}^n}\ \frac{\eta}{2} \|y-K\alpha\|^2_2 + \frac{\lambda}{2} \alpha^T K \alpha
\label{eq:}
\end{eqnarray}
Taking the derivative w.r.t. $\alpha$ provides the minimizer:
\begin{eqnarray}
\alpha^\star = (\eta K + \lambda I_n)^{-1}(\eta y)
\label{eq:}
\end{eqnarray}
Finally, unseen data points are classified as follows:
\begin{eqnarray}
x \in C_1 \textrm{ if } f^\star(x)\geq 0\\
x \in C_2 \textrm{ if } f^\star(x) < 0
\label{eq:}
\end{eqnarray}


\subsection{Laplacian-based RLS}

The Laplacian-based RLS problem for binary semi-supervised classification is as follows \cite{art:BelkinNiyogiSindhwani06SSTrans}: 
\begin{eqnarray}
\min_{f\in H_K}\ \frac{\eta}{2} \sum_{i\in n} (y_i-f_i)^2 + \frac{\lambda}{2} \|f\|^2_{H_K} + \frac{\gamma}{2} \underbrace{\sum_{i,j\in N} w_{i,j} |f_i-f_j|^2}_{\|Df\|^2},
\label{eq:LapRLS}
\end{eqnarray}
where $\|Df\|^2=\sum_{i,j\in N} w_{i,j} |f_i-f_j|^2=f^T L f$ is the Dirichlet energy and $L=D-W$ is the graph Laplacian. Observe that training data points are composed of $n$ labeled points and $N-n$ unlabeled points. Let us consider matrix $J=diag(1,...,1,0,...,0)$ with the first $n$ diagonal entries as 1 and the rest 0 and $y=[y_1,...,y_n,0,...,0]$ with $N-n$ entries as 0. This allows to write $\sum_{i\in n} (y_i-f_i)^2=\|y-Jf\|^2_2$. Representer theorem states the existence of a minimizing function $f^\star(x)=\sum_{j\in N} K(x,x_j)\alpha_j^\star$ exists. Problem \eqref{eq:LapRLS} is equivalent to
\begin{eqnarray}
\min_{\alpha\in\mathbb{R}^N}\ \frac{\eta}{2} \|y-JK\alpha\|^2_2 + \frac{\lambda}{2} \alpha^T K \alpha + \frac{\gamma}{2} (K\alpha)^T L (K\alpha).
\label{eq:}
\end{eqnarray}
Taking the derivative w.r.t. $\alpha$ provides the minimizer:
\begin{eqnarray}
\alpha^\star = (\eta JK + \lambda I_N + \gamma LK)^{-1}(\eta y)
\label{eq:}
\end{eqnarray}
Finally, unseen data points are classified as follows:
\begin{eqnarray}
x \in C_1 \textrm{ if } f^\star(x)\geq 0\\
x \in C_2 \textrm{ if } f^\star(x) < 0
\label{eq:}
\end{eqnarray}


\subsection{Total Variation-based RLS}

The TV-based RLS problem for binary semi-supervised classification is as follows \cite{art:KangShafeiSteidl12SSTrans}: 
\begin{eqnarray}
\min_{f\in H_K}\ \frac{\eta}{2} \sum_{i\in n} (y_i-f_i)^2 + \frac{\lambda}{2} \|f\|^2_{H_K} + \gamma \underbrace{\sum_{i,j\in N} w_{i,j} |f_i-f_j|}_{\|Df\|},
\label{eq:TVRLS}
\end{eqnarray}
where $\|Df\|=\sum_{i,j\in N} w_{i,j} |f_i-f_j|$ is the graph TV of function $f$. Unlike previous optimization problems, minimizing \eqref{eq:TVRLS} needs advanced optimization techniques as TV term is non-differentiable. However, recent advances in $\ell^1$ optimization provide efficient tools to deal with problem \eqref{eq:TVRLS}. In this work, we propose a splitting step coupled with an augmented Lagrangian method. Although one splitting variable is enough for minimizing \eqref{eq:TVRLS}, experimental observations suggest more accurate results using two splitting variables $g,h$. The proposed iterative optimization algorithm is as follows:
\begin{eqnarray}
(f^{n+1},h^{n+1},g^{n+1})&=&\min_{f\in H_K,h,g}\ \frac{\eta}{2} \|y-Jh\|^2_2 + \frac{\lambda}{2} \|f\|^2_{H_K} + \gamma \|Dg\| +\nonumber \\
&& <\lambda^n_1,f-g> + \frac{r_1}{2} \|f-g\|^2_2 + <\lambda^n_2,h-g> + \frac{r_2}{2} \|h-g\|^2_2 \\
\lambda^{n+1}_1&=&\lambda^n_1 + r_1(f^{n+1}-g^{n+1}) \\
\lambda^{n+1}_2&=&\lambda^n_2 + r_2(h^{n+1}-g^{n+1}) 
\label{eq:SplitTVRLS}
\end{eqnarray}

\noindent
The sub-minimization problem w.r.t. $f$ is:
\begin{eqnarray}
\min_{f\in H_K}\  \frac{\lambda}{2} \|f\|^2_{H_K} + \frac{r_1}{2} \|f-(g-\frac{\lambda_1}{r_1})\|^2_2 
\label{eq:}
\end{eqnarray}
which solution is given by $f^{n+1}=K\alpha^{n+1}$, with 
\begin{eqnarray}
\alpha^{n+1} = ( \lambda I_N + r_1 K )^{-1}( r_1 g^n - \lambda_1^n)
\label{eq:}
\end{eqnarray}

\noindent
The sub-minimization problem w.r.t. $h$ is:
\begin{eqnarray}
\min_{h}\ \frac{\eta}{2} \|y-Jh\|^2_2 + \frac{r_2}{2} \|h-(g-\frac{\lambda_2}{r_2})\|^2_2 
\label{eq:}
\end{eqnarray}
which solution is given by 
\begin{eqnarray}
h^{n+1} = ( \eta J + r_2 I_N )^{-1}( \eta y+  r_2 g^n - \lambda_2^n)
\label{eq:}
\end{eqnarray}

\noindent
The sub-minimization problem w.r.t. $g$ is:
\begin{eqnarray}
\min_{g}\ \gamma \|Dg\| + \frac{r_1}{2} \|g-(f+\frac{\lambda_1}{r_1})\|^2_2 + \frac{r_2}{2} \|g-(h+\frac{\lambda_2}{r_2})\|^2_2 
\label{eq:}
\end{eqnarray}
which can be written as
\begin{eqnarray}
\min_{g}\ \gamma \|Dg\| + \frac{r_1+r_2}{2} \|g-\frac{r_1 z_1 + r_2 z_2}{r_1+r_2}\|^2_2 
\label{eq:}
\end{eqnarray}
with $z_1=f+\frac{\lambda_1}{r_1}$ and $z_2=h+\frac{\lambda_2}{r_2}$. Different techniques can be applied to solve the TV ROF problem \cite{art:RudinOsherFatemi92ROF}. We use the primal-dual method \cite{art:ChambollePock11FastPD} which is guaranteed to converge in $O(\frac{1}{k^2})$, $k$ being the iteration number. Finally, we project each function $f,h,g$ on the unit ball (i.e. $f^{n+1}\leftarrow N. \frac{f^{n+1}}{\|f^{n+1}\|_2}$) and constraint them to be zero-mean (i.e. $f^{n+1}\leftarrow f^{n+1} - mean(f^{n+1})$).\\

\noindent
We summarize the iterative algorithm:
\begin{eqnarray}
\alpha^{n+1} &=& ( \lambda I_N + r_1 K )^{-1}( r_1 g^n - \lambda_1^n)\\
f^{n+1}&=&K\alpha^{n+1}\\
h^{n+1} &=& ( \eta J + r_2 I_N )^{-1}( \eta y +  r_2 g^n - \lambda_2^n)\\
\bar{g}^{n+1} &=&\argmin_{g}\ \gamma \|Dg\| + \frac{r_1+r_2}{2} \|g-\frac{r_1 z_1 + r_2 z_2}{r_1+r_2}\|^2_2\\
&& \textrm{ with } z_1=f+\frac{\lambda_1^n}{r_1},\ z_2=h+\frac{\lambda_2^n}{r_2}\\
\hat{g}^{n+1} &=& N. \frac{\bar{g}^{n+1}}{\|\bar{g}^{n+1}\|_2}\\
g^{n+1} &=& \hat{g}^{n+1} - mean(\hat{g}^{n+1})
\label{eq:AlgoTVRLS}
\end{eqnarray}


\subsection{Cheeger-based RLS}

The Cheeger-based RLS problem for binary semi-supervised classification is as follows: 
\begin{eqnarray}
\min_{f\in H_K}\ \frac{\sum_{i,j\in N} w_{i,j} |f_i-f_j|}{\sum_{i\in N}|f_i-median(f)|}\ s.t.\ f_i=y_i, \forall i\in n
\label{eq:CheegerRLS}
\end{eqnarray}

\noindent
Based on \cite{art:BressonLaurentUminskyVonBrecht12CheegerConverg}, the following algorithm is proposed:
\begin{eqnarray}
g^{n+1} &=& f^n + c.sign(f^n)\\
e^{n+1} &=&\textrm{RLS}(g^{n+1})\\
h^{n+1} &=& \argmin_{h} \ TV(h) + \frac{E^n}{2c} \|h-e^{n+1}\|^2_2\\
t^{n+1} &=& h^{n+1} - median(h^{n+1})\\
s^{n+1} &=& 
\left\{
\begin{array}{lll}
y_i & \forall i\in n\\
t^{n+1}(i) & \forall i\not\in n
\end{array}
\right.
\\
f^{n+1} &=& N.\frac{s^{n+1}}{\|s^{n+1}\|_2}
\label{eq:AlgoCheegerRLS}
\end{eqnarray}
where RLS($g$) is as follow
\begin{eqnarray}
\begin{aligned}
\min_{e\in H_K} \frac{\lambda}{2}||e||^2_{H_K}+\frac{r}{2}||e-g||^2_2,
\label{eq:CheegerRLS2}
\end{aligned}
\end{eqnarray}
which solution is given by $e^{n+1}=K\alpha^*$, with
\begin{eqnarray}
\alpha^*=(\lambda I+rK)^{-1}rg.
\end{eqnarray}


\subsection{Support Vector Machine (SVM)}

The standard SVM method for binary classification is as follows \cite{art:CortesVapnik95SVM}. Find a function $f:\mathbb{R}^d\rightarrow\mathbb{R}$ such that
\begin{eqnarray}
\begin{aligned}
\min_{f\in H_K,b\in\mathbb{R}} & \ \frac{\lambda}{2}||f||_{H_K}^2,\\
\textrm{s.t.} & \  y_i(f_i+b)\geq 1, i=1,\dots,n.
\label{eq:SVM}
\end{aligned}
\end{eqnarray}
where $\lambda>0$. To deal with non-separable case, the above problem can be rewritten with a slack variable $\xi$:
\begin{eqnarray}
\begin{aligned}
\min_{f\in H_K,b\in\mathbb{R},\xi\in\mathbb{R}^n} & \ \frac{\lambda}{2}||f||_{H_K}^2+\mu\sum_{i\in n}\xi_i,\\
\textrm{s.t.} & \  y_i(f_i+b)\geq 1-\xi_i, i=1,\dots,n,\\
& \ \xi_i\geq 0, i=1,\dots,n
\label{eq:CSVM}
\end{aligned}
\end{eqnarray}
Representer theorem states the existence of a minimizing function $f^\star(x)=\sum_{j\in n} K(x,x_j)\alpha_j^\star$ and $\|f\|^2_{H_K}=\alpha^T K \alpha$. Problem \eqref{eq:CSVM} is equivalent to
\begin{eqnarray}
\begin{aligned}
\min_{\alpha,\xi\in\mathbb{R}^n,b\in\mathbb{R}} & \ \frac{\lambda}{2}\alpha^TK\alpha+\mu\sum_{i=1}^n\xi_i,\\
\textrm{s.t.} & \  y_i(\sum_{j=1}^nK(x_i,x_j)\alpha_j+b)\geq 1-\xi_i,i=1,\dots,n,\\
& \ \xi_i\geq 0, i=1,\dots,n
\label{eq:CSVM2}
\end{aligned}
\end{eqnarray}
By using the Lagrangian multiplier technique, problem (\ref{eq:CSVM2}) can be reformulated as:
\begin{eqnarray}
\begin{aligned}
\min_{\alpha,\xi,\beta,\beta_\xi\in\mathbb{R}^n,b\in\mathbb{R}} & \ \frac{\lambda}{2}\alpha^TK\alpha+\mu\xi^T\bold{1}+\beta^T(\bold{1}-\xi-Y(K\alpha+b\bold{1}))-\beta_\xi^T\xi,\\
\textrm{s.t.} & \ \beta_i, \beta_{\xi i}\geq 0,i=1,\dots,n
\label{eq:CSVM3}
\end{aligned}
\end{eqnarray}
where $\beta, \beta_\xi$ are Lagrangian multipliers, $\bold{1}$ is a vector whose elements are all ones, and $Y=diag(y_1,...,y_n)$. Let us consider the Lagrangian optimality conditions. Taking the derivative w.r.t. $b$ and setting to $0$ gives
\begin{eqnarray}
\-\beta^TY\bold{1}=0 \ \Rightarrow \ \beta^Ty=0.
\label{eq:b}
\end{eqnarray}
Taking the derivative w.r.t. $\xi$ and setting to $0$ gives
\begin{eqnarray}
\mu\bold{1}-\beta-\beta_\xi=0 \ \Rightarrow \ 0\leq \beta_i\leq \mu, i=1,\dots,n.
\label{eq:xi}
\end{eqnarray}
Taking the derivative w.r.t. $\alpha$ and setting to $0$ gives
\begin{eqnarray}
\alpha=\frac{Y\beta}{\lambda}
\label{eq:alpha}
\end{eqnarray}
By substituting (\ref{eq:alpha}) back into (\ref{eq:CSVM3}), we reach the following dual optimization problem:
\begin{eqnarray}
\begin{aligned}
\max_{\beta\in\mathbb{R}^n} & \ \beta^T\bold{1}-\frac{1}{2}\beta^TQ\beta,\\
\textrm{s.t.} & \ \beta^Ty=0,\\
& \ 0\leq\beta_i\leq \mu,i=1,\dots,n
\label{eq:SVMD}
\end{aligned}
\end{eqnarray}
where $Q=Y(\frac{K}{\lambda})Y$.  The above problem can be solved using several efficient SVM solvers s.a. libSVM \cite{art:ChangLin11LibSVM}. Once the optimal $\beta^*$ is obtained, it is straightforward to get the optimal $\alpha^*$:
\begin{eqnarray}
\alpha^*=\frac{Y\beta^*}{\lambda}
\end{eqnarray}
and
\begin{eqnarray}
f^*(x)=\sum_{i=1}^n\alpha_i^*K(x,x_i).
\end{eqnarray}
Finally, unseen data points are classified as follows:
\begin{eqnarray}
x \in C_1 \textrm{ if } f^\star(x)\geq 0\\
x \in C_2 \textrm{ if } f^\star(x) < 0
\label{eq:}
\end{eqnarray}


\subsection{Laplacian-based SVM}

The Laplacian-based SVM problem with slack variable for binary semi-supervised classification is as follows \cite{art:BelkinNiyogiSindhwani06SSTrans}: 
\begin{eqnarray}
\begin{aligned}
\min_{f\in H_K,\xi\in\mathbb{R}^N,b\in\mathbb{R}} & \ \frac{\lambda}{2} \|f\|^2_{H_K} + \mu\sum_{i\in N}\xi_i + \frac{\gamma}{2} \underbrace{\sum_{i,j\in N} w_{i,j} |f_i-f_j|^2}_{\|Df\|^2},\\
\textrm{s.t.} & \ y_i(f_i+b)\geq 1-\xi_i, i=1,\dots,N\\
& \ \xi_i\geq 0, i=1,\dots,N
\label{eq:LapSVM}
\end{aligned}
\end{eqnarray}
By using Lagrangian multipliers technique, problem (\ref{eq:LapSVM}) becomes:
\begin{eqnarray}
\begin{aligned}
\min_{\alpha,\xi,\beta,\beta_\xi\in\mathbb{R}^N,b\in\mathbb{R}} & \ \frac{\lambda}{2}\alpha^TK\alpha+\mu\xi^T\bold{1}+\frac{\gamma}{2}\alpha^TKLK\alpha+\beta^T(\bold{1}-\xi-Y(K\alpha+b\bold{1}))-\beta_\xi^T\xi,\\
\textrm{s.t.} & \ \beta_i, \beta_{\xi i}\geq 0,i=1,\dots,N
\label{eq:LapSVM2}
\end{aligned}
\end{eqnarray}
Applying the same steps as (\ref{eq:b}),(\ref{eq:xi}) and (\ref{eq:alpha}), we get
\begin{eqnarray}
\begin{aligned}
\max_{\beta\in\mathbb{R}^N} & \ \beta^T\bold{1}-\frac{1}{2}\beta^TQ\beta,\\
\textrm{s.t.} & \ \beta^Ty=0,\\
& \ 0\leq\beta_i\leq \mu,i=1,\dots,N
\label{eq:SVMD}
\end{aligned}
\end{eqnarray}
where 
\begin{eqnarray}
Q=Y(\lambda I+\gamma LK)^{-1}KY
\end{eqnarray}
Optimal $\alpha^*$ is obtained by solving the following linear system:
\begin{eqnarray}
\alpha^*=(\lambda I+\gamma LK)^{-1}Y\beta^*
\label{alpha1}
\end{eqnarray}
and
\begin{eqnarray}
f^*(x)=\sum_{i=1}^N\alpha_i^*K(x,x_i).
\end{eqnarray}
Finally, unseen data points are classified as follows:
\begin{eqnarray}
x \in C_1 \textrm{ if } f^\star(x)\geq 0\\
x \in C_2 \textrm{ if } f^\star(x) < 0
\label{eq:}
\end{eqnarray}


\subsection{Total Variation-based SVM}

The TV-based SVM for binary semi-supervised classification is as follows: 
\begin{eqnarray}
\begin{aligned}
\min_{f\in H_K,\xi\in\mathbb{R}^N,b\in\mathbb{R}} & \ \frac{\lambda}{2} \|f\|^2_{H_K} + \mu\sum_{i\in N}\xi_i + \gamma \underbrace{\sum_{i,j\in N} w_{i,j} |f_i-f_j|}_{\|Df\|},\\
\textrm{s.t.} & \ y_i(f_i+b)\geq 1-\xi_i, i=1,\dots,N\\
& \ \xi_i\geq 0, i=1,\dots,N
\label{eq:TVSVM}
\end{aligned}
\end{eqnarray}
where $\|Df\|=\sum_{i,j\in N} w_{i,j} |f_i-f_j|$ is the graph TV of function $f$. Like for TV-based RLS, we use  a splitting step coupled with an augmented Lagrangian method. The proposed iterative optimization algorithm is as follows:
\begin{eqnarray}
(f^{n+1},h^{n+1},g^{n+1})&=&\min_{f\in H_K,h,g,\xi,b}\ \frac{\lambda}{2} \|f\|^2_{H_K} + \mu\sum_{i=1}^N\xi_i +  \gamma \|Dg\|+ \nonumber\\ && <\lambda^n_1,f-g> + \frac{r_1}{2} \|f-g\|^2_2 + <\lambda^n_2,h-g> + \frac{r_2}{2} \|h-g\|^2_2\\
\textrm{s.t.} & & \ y_i(h_i+b)\geq 1-\xi_i, i=1,\dots,N\\
& & \ \xi_i\geq 0, i=1,\dots,N \nonumber \\
&&  \\
\lambda^{n+1}_1&=&\lambda^n_1 + r_1(f^{n+1}-g^{n+1}) \\
\lambda^{n+1}_2&=&\lambda^n_2 + r_2(h^{n+1}-g^{n+1}) 
\label{eq:SplitTVRLS}
\end{eqnarray}

\noindent
The sub-minimization problem w.r.t. $f$ is:
\begin{eqnarray}
\min_{f\in H_K}\  \frac{\lambda}{2} \|f\|^2_{H_K} + \frac{r_1}{2} \|f-(g-\frac{\lambda_1}{r_1})\|^2_2 
\label{eq:}
\end{eqnarray}
which solution is given by $f^{n+1}=K\alpha^{n+1}$, with 
\begin{eqnarray}
\alpha^{n+1} = ( \lambda I_N + r_1 K )^{-1}( r_1 g^n - \lambda_1^n)
\label{eq:}
\end{eqnarray}

\noindent
The sub-minimization problem w.r.t. $h$ is:
\begin{eqnarray}
\begin{aligned}
\min_{h\xi,b} & \ \mu\sum_{i=1}^N\xi_i+\frac{r_2}{2} \|h-e\|^2_2 \\
\textrm{s.t.} & \ y_i(h_i+b)\geq 1-\xi_i, i=1,\dots,N\\
& \ \xi_i\geq 0, i=1,\dots,N 
\label{eq:SplitTVSVM}
\end{aligned}
\end{eqnarray}
where $e=g-\frac{\lambda_2}{r_2}$. Problem (\ref{eq:SplitTVSVM}) is equivalent to
\begin{eqnarray}
\begin{aligned}
\min_{h\xi,b,\beta,\beta_\xi} & \ \mu\xi^T\bold{1}+\frac{r_2}{2} \|h-e\|^2_2+\beta^T(\bold{1}-\xi-Y(h+b))-\beta_{\xi}^T\xi \\
\textrm{s.t.} & \ \beta_i, \beta_{\xi i}\geq 0, i=1,\dots,N 
\label{eq:SplitTVSVM2}
\end{aligned}
\end{eqnarray}

\noindent
Applying the same steps as (\ref{eq:b}),(\ref{eq:xi}) and (\ref{eq:alpha}) did, we get
\begin{eqnarray}
\begin{aligned}
\max_{\beta\in\mathbb{R}^N} & \ \beta^T\bold{1}-\frac{1}{2}\beta^TQ\beta-\beta^TP,\\
\textrm{s.t.} & \ \beta^Ty=0,\\
& \ 0\leq\beta_i\leq \mu,i=1,\dots,N
\label{eq:TVSVMD1}
\end{aligned}
\end{eqnarray}
where $Q=\frac{YY}{r_2}$ and $P=Ye$. Problem (\ref{eq:TVSVMD1}) can be solved by gradient descent method, and the solution $\beta^*$ can be used to obtain the optimal $h^{n+1}$:
\begin{eqnarray}
h^{n+1} = \frac{1}{r_2}Y\beta^*+e
\label{alpha2}
\end{eqnarray}

\noindent
The sub-minimization problem w.r.t. $g$ is:
\begin{eqnarray}
\min_{g}\ \gamma \|Dg\| + \frac{r_1}{2} \|g-(f+\frac{\lambda_1}{r_1})\|^2_2 + \frac{r_2}{2} \|g-(h+\frac{\lambda_2}{r_2})\|^2_2 
\label{eq:}
\end{eqnarray}
which can be written as
\begin{eqnarray}
\min_{g}\ \gamma \|Dg\| + \frac{r_1+r_2}{2} \|g-\frac{r_1 z_1 + r_2 z_2}{r_1+r_2}\|^2_2 
\label{eq:}
\end{eqnarray}
with $z_1=f+\frac{\lambda_1}{r_1}$ and $z_2=h+\frac{\lambda_2}{r_2}$. 

\noindent
We summarize the iterative algorithm:
\begin{eqnarray}
\alpha^{n+1} &=& ( \lambda I_N + r_1 K )^{-1}( r_1 g^n - \lambda_1^n)\\
f^{n+1}&=&K\alpha^{n+1}\\
\beta^* &=&\max_{\beta\in\mathbb{R}^N} \beta^T\bold{1}-\frac{1}{2}\beta^TQ\beta-\beta^TP,\ \textrm{s.t.}\ \beta^Ty=0,0\leq\beta_i\leq C,i=1,\dots,N\\
h^{n+1} &=& \frac{1}{r_2}Y\beta^*+e\\
\bar{g}^{n+1} &=&\argmin_{g}\ \gamma \|Dg\| + \frac{r_1+r_2}{2} \|g-\frac{r_1 z_1 + r_2 z_2}{r_1+r_2}\|^2_2\\
&& \textrm{ with } z_1=f+\frac{\lambda_1^n}{r_1},\ z_2=h+\frac{\lambda_2^n}{r_2}\\
\hat{g}^{n+1} &=& N. \frac{\bar{g}^{n+1}}{\|\bar{g}^{n+1}\|_2}\\
g^{n+1} &=& \hat{g}^{n+1} - mean(\hat{g}^{n+1})
\label{eq:AlgoTVRLS}
\end{eqnarray}


\subsection{Cheeger-based SVM}

The Cheeger-based SVM problem for binary semi-supervised classification is as follows: 
\begin{eqnarray}
\min_{f\in H_K}\ \frac{\sum_{i,j\in N} w_{i,j} |f_i-f_j|}{\sum_{i\in N}|f_i-median(f)|}\ s.t.\ f_i=y_i,\ \forall i\in n\\
\textrm{s.t.} \ y_i(f_i+b)\geq 1-\xi_i, i=1,\dots,N\\
 \ \xi_i\geq 0, i=1,\dots,N
\label{eq:CheegerRLS}
\end{eqnarray}

\noindent
Based on \cite{art:BressonLaurentUminskyVonBrecht12CheegerConverg}, the following algorithm is proposed:
\begin{eqnarray}
g^{n+1} &=& f^n + c.sign(f^n)\\
e^{n+1} &=& \textrm{SVM}(g^{n+1})\\
h^{n+1} &=& \argmin_{h} \ TV(h) + \frac{E^n}{2c} \|h-e^{n+1}\|^2_2\\
t^{n+1} &=& h^{n+1} - median(h^{n+1})\\
s^{n+1} &=& 
\left\{
\begin{array}{lll}
l(i) & \forall i\in n\\
t^{n+1}(i) & \forall i\not\in n
\end{array}
\right.
\\
f^{n+1} &=& N.\frac{s^{n+1}}{\|s^{n+1}\|_2}
\label{eq:AlgoCheegerRLS}
\end{eqnarray}
where SVM$(g)$ is as follow:
\begin{eqnarray}
\begin{aligned}
\min_{e,\xi,b} & \ \frac{\lambda}{2}||e||_{H_K}^2+\mu\sum_{i\in N}\xi_i+\frac{r}{2} \|e-g\|^2_2 \\
\textrm{s.t.} & \ y_i(e_i+b)\geq 1-\xi_i, i=1,\dots,N\\
& \ \xi_i\geq 0, i=1,\dots,N 
\label{eq:CheegerSVM}
\end{aligned}
\end{eqnarray}
Problem (\ref{eq:CheegerSVM}) is equivalent to
\begin{eqnarray}
\begin{aligned}
\min_{e,\xi,b,\beta,\beta_\xi} & \ \frac{\lambda}{2}||e||_{H_K}^2+\mu\xi^T\bold{1}+\frac{r}{2} \|e-g\|^2_2+\beta^T(\bold{1}-\xi-Y(e+b))-\beta_{\xi}^T\xi \\
\textrm{s.t.} & \ \beta_i, \beta_{\xi i}\geq 0, i=1,\dots,N 
\label{eq:CheegerSVM2}
\end{aligned}
\end{eqnarray}
Applying the same steps as (\ref{eq:b}),(\ref{eq:xi}) and (\ref{eq:alpha}), we get
\begin{eqnarray}
\begin{aligned}
\max_{\beta\in\mathbb{R}^N} & \ \beta^T\bold{1}-\frac{1}{2}\beta^TQ\beta-\frac{1}{2}P\beta,\\
\textrm{s.t.} & \ \beta^Ty=0,\\
& \ 0\leq\beta_i\leq \mu,i=1,\dots,N
\label{eq:TVSVMD}
\end{aligned}
\end{eqnarray}
where $Q=YGY$, $P=rg^T(G+G^T)Y$ and $G=(\lambda I+rK)^{-1}K$. The above problem can be solved by gradient descent method, and the solution $\beta^*$ can be used to obtain the optimal $\alpha^*$:
\begin{eqnarray}
\alpha^*=(\lambda I+rK)^{-1}(Y\beta^*+rg)
\label{alpha3}
\end{eqnarray}
and
\begin{eqnarray}
e^{n+1}=K\alpha^*
\label{eq:}
\end{eqnarray}


\subsection{Experimental results}

\begin{table}[h!]
\vspace{0.5cm}
\centering
\begin{tabular}{|c|c|c|c|c|c|c|c|c|c|}
\hline	
\# labels per class &  1 & 5 & 10 & 50  \\
\hline	
Lap-RLS        &  18.09 & 10.48 & 7.77 & 4.14 \\
\hline 
Lap-SVM       &  13.79 & 9.84   & 7.61 & 4.77\\
\hline	
TV-RLS         & {\bf 3.18}  & 3.16    & {\bf 3.13} & 3.16 \\
\hline
TV-SVM        & {\bf 3.18}  & {\bf 3.13}    & {\bf 3.13 }& 3.08\\
\hline	
Cheeger-RLS  & 4.06  & 3.74    & 4.03 &  2.84\\
\hline
Cheeger-SVM & 3.87  & 3.74    & 4.00 & {\bf 2.73 }\\
\hline	
\end{tabular}
\caption{Binary semi-supervised classification algorithms tested on the sets of 4's and 9's from USPS dataset. Error is averaged over 10 runs with randomly selected labels.}
\label{tab:}
\end{table}



\section{Multi-class data classification}




\subsection{Laplacian-based RLS}

The Laplacian-based RLS problem for multi-class semi-supervised classification is as follows: 
\begin{eqnarray}
\min_{\vec{f}=(f^1,...,f^c)\in H_K}\ \frac{\eta}{2} \sum_{k=1}^c \sum_{i\in n} (y^k_i-f^k_i)^2 + \frac{\lambda}{2} \sum_{k=1}^c \|f^k\|^2_{H_K} + \frac{\gamma}{2} \sum_{k=1}^c \underbrace{\sum_{i,j\in N} w_{i,j} |f^k_i-f^k_j|^2}_{\|Df^k\|^2},\nonumber\\
\textrm{s.t. } \sum_{k=1}^c f^k_i=1,\ f^k_i\geq 0, \forall i\in N
\label{eq:LapRLSMC}
\end{eqnarray}
where the last constraint being the simplex constraint. Problem \eqref{eq:LapRLSMC} is equivalent to
\begin{eqnarray}
\min_{\vec{f}=(f^1,...,f^c)\in H_K}\ \frac{\eta}{2} \sum_{k=1}^c \sum_{i\in n} (y^k_i-f^k_i)^2 + \frac{\lambda}{2} \sum_{k=1}^c \|f^k\|^2_{H_K} + \frac{\gamma}{2} \sum_{k=1}^c \|Df^k\|^2,\nonumber\\
\textrm{s.t. } f^k=g^k,\ \sum_{k=1}^c g^k_i=1,\ g^k_i\geq 0, \forall i\in N
\label{eq:LapRLSMC2}
\end{eqnarray}
This leads to the proposed iterative algorithm:
\begin{eqnarray}
(\alpha^k)^{n+1}&=&\argmin_{\alpha^k\in\mathbb{R}^N}\ \frac{\eta}{2} \|y^k-J^kK\alpha^k\|^2_2 + \frac{\lambda}{2} \alpha_k^T K \alpha_k + \frac{\gamma}{2} (K\alpha_k)^T L (K\alpha_k) \nonumber\\
&&+ \frac{r}{2} \|K\alpha^k-(g^k-\frac{\lambda^k}{r})\|^2_2\\
&=& (\eta J^kK + rK + \lambda I_N + \gamma LK)^{-1}(\eta y^k+rg^k-\lambda^k)\\
(f^k)^{n+1}&=&K(\alpha^k)^{n+1}\\
(g^k)^{n+1}&=&\Pi_{\sum g^k=1}(f^k+\frac{\lambda^k}{r})
\label{eq:}
\end{eqnarray}
The simplex projection is done by Michelot's method \cite{michelot1986finite}.

\noindent
Finally, unseen data points are classified as follows:
\begin{eqnarray}
x \in C_k \textrm{ if } f^\star_k(x) = \max_j (\{f^\star_j(x)\}_{1\leq j \leq c})
\label{eq:}
\end{eqnarray}


\subsection{Total Variation-based RLS}

The TV-based RLS problem for multi-class semi-supervised classification is as follows: 
\begin{eqnarray}
\min_{\vec{f}=(f^1,...,f^c)\in H_K}\ \frac{\eta}{2} \sum_{k=1}^c \sum_{i\in n} (y^k_i-f^k_i)^2 + \frac{\lambda}{2} \sum_{k=1}^c \|f^k\|^2_{H_K} + \gamma \underbrace{\sum_{i,j\in N} w_{i,j} |f^k_i-f^k_j|}_{\|Df^k\|},\nonumber\\
\textrm{s.t. } \sum_{k=1}^c f^k(i)=1,\ f^k(i)\geq 0, \forall i\in N
\label{eq:TVRLSMC}
\end{eqnarray}
Problem \eqref{eq:TVRLSMC} is equivalent to
\begin{eqnarray}
\min_{\vec{f}=(f^1,...,f^c)\in H_K}\ \frac{\eta}{2} \sum_{k=1}^c \sum_{i\in L} (y^k_i-f^k_i)^2 + \frac{\lambda}{2} \sum_{k=1}^c \|f^k\|^2_{H_K} + \frac{\gamma}{2} \sum_{k=1}^c \|Df^k\|,\nonumber\\
\textrm{s.t. } f^k=g^k,\ \sum_{k=1}^c g^k_i=1,\ g^k_i\geq 0, \forall i\in N
\label{eq:TVRLSMC2}
\end{eqnarray}
This leads to the proposed iterative algorithm:
\begin{eqnarray}
(\alpha^k)^{n+1}&=&\argmin_{\alpha^k\in\mathbb{R}^N}\ \frac{\eta}{2} \|y^k-J^kK\alpha^k\|^2_2 + \frac{\lambda}{2} \alpha_k^T K \alpha_k + \frac{r}{2} \|K\alpha^k-(g^k-\frac{\lambda^k}{r})\|^2_2\\
&=& (\eta J^kK + rK + \lambda I_N )^{-1}(\eta y^k+rg^k-\lambda^k)\nonumber\\
(f^k)^{n+1}&=&K(\alpha^k)^{n+1}\\
(\hat{g}^k)^{n+1}&=&\argmin_{g^k}\ \gamma \|Dg^k\| + \frac{r}{2} \|g^k-(f^k+\frac{\lambda^k}{r})\|^2_2\\
(\bar{g}^k)^{n+1}&=&\Pi_{\sum g^k=1}(\hat{g}^k)\\
(g^k)^{n+1}&=& N.\frac{(\bar{g}^k)^{n+1}}{\|(\bar{g}^k)^{n+1}\|_2}
\label{eq:}
\end{eqnarray}

\noindent
Finally, unseen data points are classified as follows:
\begin{eqnarray}
x \in C_k \textrm{ if } f^\star_k(x) = \max_j (\{f^\star_j(x)\}_{1\leq j \leq c})
\label{eq:}
\end{eqnarray}


\subsection{Cheeger-based RLS}

The Cheeger-based RLS problem for multi-class semi-supervised classification is as follows: 
\begin{eqnarray}
\min_{\vec{f}=(f^1,...,f^c)\in H_K}\ \sum_{k=1}^c \frac{\sum_{i,j\in N} w_{i,j} |f^k_i-f^k_j|}{\sum_{i\in N}|f^k_i-median(f^k)|}\ s.t.\ f^k_i=l^k_i,\ \forall i\in n\\
\label{eq:CheegerRLSMC}
\end{eqnarray}
The following algorithm is proposed:
\begin{eqnarray}
(g^k)^{n+1} &=& (f^k)^n + c.sign((f^k)^n)\\
 (e^k)^{n+1}&=& \textrm{RLS}((g^k)^{n+1})\\
(h^k)^{n+1} &=& \argmin_{h^k} \ TV(h^k) + \frac{E^n}{2c} \|h^k-(e^k)^{n+1}\|^2_2\\
(t^k)^{n+1} &=& (h^k)^{n+1} - median((h^k)^{n+1})\\
(s^k)^{n+1} &=& 
\left\{
\begin{array}{lll}
y^k_i & \forall i\in n\\
(t^k)^{n+1}(i) & \forall i\not\in n
\end{array}
\right.
\\
(\hat{s}^k)^{n+1}&=&\Pi_{\sum s^k=1}(s^k)\\
(f^k)^{n+1} &=& N.\frac{(\hat{s}^k)^{n+1}}{\|(\hat{s}^k)^{n+1}\|_2}
\label{eq:AlgoCheegerRLSMC}
\end{eqnarray}
where RLS($g$) is exact the same as (\ref{eq:CheegerRLS2}).

\noindent
Finally, unseen data points are classified as follows:
\begin{eqnarray}
x \in C_k \textrm{ if } f^\star_k(x) = \max_j (\{f^\star_j(x)\}_{1\leq j \leq c})
\label{eq:}
\end{eqnarray}

\subsection{Laplacian-based SVM}
The Laplacian-based SVM for multi-class semi-supervised classification is as follows: 
\begin{eqnarray}
\begin{aligned}
\min_{\vec{f}=(f^1,...,f^c)\in H_K, b\in\mathbb{R}^c, \xi\in\mathbb{R}^{N\times c}} & \ \frac{\lambda}{2} \sum_{k=1}^c \|f^k\|^2_{H_K} + \mu\sum_{k=1}^c\sum_{i\in N}\xi_i^k+\frac{\gamma}{2} \sum_{k=1}^c \underbrace{\sum_{i,j\in N} w_{i,j} |f^k_i-f^k_j|^2}_{\|Df^k\|^2},\nonumber\\
\textrm{s.t. } & \  y_i^k(f_i^k+b^k)\geq 1-\xi_i^k,\xi_i^k\geq 0,i\in N, k\in c\\
& \ \sum_{k=1}^c f^k_i=1,\ f^k(i)\geq 0, \forall i\in N\\
\label{eq:LapSVMMC}
\end{aligned}
\end{eqnarray}
Problem \eqref{eq:LapSVMMC} is equivalent to
\begin{eqnarray}
\begin{aligned}
\min_{\vec{f}=(f^1,...,f^c)\in H_K, b\in\mathbb{R}^c,\xi\in\mathbb{R}^{N\times c}} & \ \frac{\lambda}{2} \sum_{k=1}^c \|f^k\|^2_{H_K} + \mu\sum_{k=1}^c\sum_{i\in N}\xi_i^k+\frac{\gamma}{2} \sum_{k=1}^c \|Df^k\|^2,\nonumber\\
\textrm{s.t. } & \  y_i^k(f_i^k+b^k)\geq 1-\xi_i^k,\xi_i^k\geq 0, i\in N, k\in c\\
& \  f^k=g^k,\ \sum_{k=1}^c g_i^k=1,\ g_i^k\geq 0, \forall i\in N
\label{eq:LapSVMMC2}
\end{aligned}
\end{eqnarray}
Notes that, each $f^k$ can be solved independently by using the same procedure as below (superscript $k$ is ignored for convenience):
\begin{eqnarray}
\begin{aligned}
\min_{f\in H_K,b\in \mathbb{R},\xi\in\mathbb{R}^N} & \ \frac{\lambda}{2}||f||_{H_K}+\mu\xi^T\bold{1}+\frac{\gamma}{2}f^TLf+\frac{r}{2}||f-e||_2^2,\\
\textrm{s.t.} & \ y_i(f_i+b)\geq 1-\xi_i,\xi_i\geq 0, i\in N
\end{aligned}
\label{eq:LapSVMOC} 
\end{eqnarray}
where $e=g-\frac{l}{r}$, and $l$ is the Lagrangian multiplier. Problem (\ref{eq:LapSVMOC}) is equivalent to
\begin{eqnarray}
\begin{aligned}
\min_{b\in \mathbb{R},\alpha,\xi,\beta,\beta_\xi\in\mathbb{R}^N} & \ \frac{\lambda}{2}\alpha^TK\alpha+\mu\xi^T\bold{1}+\frac{\gamma}{2}\alpha^TKLK\alpha+\frac{r}{2}||K\alpha-e||_2^2\\
& \ +\beta^T(\bold{1}-\xi-Y(K\alpha+b))-\beta_{\xi}^T\xi\\
\textrm{s.t.} & \ \beta,\beta_\xi\geq 0, i\in N
\end{aligned}
\label{eq:LapSVMOC2} 
\end{eqnarray}
Applying the same steps as (\ref{eq:b}),(\ref{eq:xi}) and (\ref{eq:alpha}), we get
\begin{eqnarray}
\begin{aligned}
\max_{\beta\in\mathbb{R}^N} & \ \beta^T\bold{1}-\frac{1}{2}\beta^TQ\beta-\frac{1}{2}P\beta,\\
\textrm{s.t.} & \ \beta^Ty=0,\\
& \ 0\leq\beta_i\leq \mu,i=1,\dots,N
\label{eq:LapSVMOCD}
\end{aligned}
\end{eqnarray}
where $Q=YGY$, $P=re^T(G+G^T)Y$ and $G=(\lambda I+\gamma LK+rK)^{-1}K$. The above problem can be solved by gradient descent method, and the solution $\beta^*$ can be used to obtain the optimal $\alpha^*$, which is:
\begin{eqnarray}
\alpha^*=(\lambda I+\gamma LK+rK)^{-1}(Y\beta^*+re)
\label{eq:newalpha}
\end{eqnarray}
and
\begin{eqnarray}
f=K\alpha^*
\label{eq:newf}
\end{eqnarray}
This leads to the following iterative algorithm:
\begin{eqnarray}
(f^k)^{n+1}&=&\textrm{computed by using (\ref{eq:LapSVMOCD}), (\ref{eq:newalpha}) and (\ref{eq:newf})}\\
(g^k)^{n+1}&=& \Pi_{\sum g^k=1}((f^k)^{n+1}+\frac{l^k}{r}).
\end{eqnarray}
The simplex projection is done by Michelot's method \cite{michelot1986finite}.

\noindent
Finally, unseen data points are classified as follows:
\begin{eqnarray}
x \in C_k \textrm{ if } f^\star_k(x) = \max_j (\{f^\star_j(x)\}_{1\leq j \leq c})
\label{eq:}
\end{eqnarray}


\subsection{Total Variation-based SVM}

The TV-based SVM for multi-class semi-supervised classification is as follows: 
\begin{eqnarray}
\begin{aligned}
\min_{\vec{f}=(f^1,...,f^c) \ \in H_K,\xi\in\mathbb{R}^{N\times c},b\in\mathbb{R}^c} & \ \frac{\lambda}{2} \sum_{k=1}^c\|f^k\|^2_{H_K} + \mu\sum_{k=1}^c\sum_{i=1}^N\xi_i^k + \gamma \sum_{k=1}^c\underbrace{\sum_{i,j\in N} w_{i,j} |f_i^k-f_j^k|}_{\|Df^k\|},\\
\textrm{s.t.} & \ y_i^k(f_i^k+b^k)\geq 1-\xi_i^k,\xi_i^k\geq 0, i\in N,k\in c
\label{eq:TVSVMMC}
\end{aligned}
\end{eqnarray}
Problem \eqref{eq:TVSVMMC} is equivalent to
\begin{eqnarray}
\begin{aligned}
\min_{\vec{f}=(f^1,...,f^c) \ \in H_K,\xi\in\mathbb{R}^{N\times c},b\in\mathbb{R}^c} & \ \frac{\lambda}{2} \sum_{k=1}^c\|f^k\|^2_{H_K} + \mu\sum_{k=1}^c\sum_{i=1}^N\xi_i^k + \gamma \sum_{k=1}^c\|Dg^k\|,\\
\textrm{s.t.} & \ y_i^k(f_i^k+b^k)\geq 1-\xi_i^k,\xi_i^k\geq 0, i\in N,k\in c\\
& \  f^k=g^k,\ \sum_{k=1}^c g^k(i)=1,\ g^k(i)\geq 0, \forall i\in N
\label{eq:TVRLSMC2}
\end{aligned}
\end{eqnarray}
Notes that, each $f^k$ can be solved independently:
\begin{eqnarray}
\begin{aligned}
\min_{f\in H_K,b\in \mathbb{R},\xi\in\mathbb{R}^N} & \ \frac{\lambda}{2}||f||_{H_K}+\mu\xi^T\bold{1}+\frac{r}{2}||f-e||_2^2,\\
\textrm{s.t.} & \ y_i(f_i+b)\geq 1-\xi_i,\xi_i\geq 0, i\in N
\end{aligned}
\label{eq:TVSVMOC} 
\end{eqnarray}
where $e=g-\frac{l}{r}$, and $l$ is the Lagrangian multiplier. Problem (\ref{eq:TVSVMOC}) is equivalent to
\begin{eqnarray}
\begin{aligned}
\min_{b\in \mathbb{R},\alpha,\xi,\beta,\beta_\xi\in\mathbb{R}^N} & \ \frac{\lambda}{2}\alpha^TK\alpha+\mu\xi^T\bold{1}+\frac{r}{2}||K\alpha-e||_2^2+\beta^T(\bold{1}-\xi-Y(K\alpha+b))-\beta_{\xi}^T\xi\\
\textrm{s.t.} & \ \beta,\beta_\xi\geq 0, i\in N
\end{aligned}
\label{eq:LapSVMOC2} 
\end{eqnarray}
Applying the same steps as (\ref{eq:b}),(\ref{eq:xi}) and (\ref{eq:alpha}), we get
\begin{eqnarray}
\begin{aligned}
\max_{\beta\in\mathbb{R}^N} & \ \beta^T\bold{1}-\frac{1}{2}\beta^TQ\beta-\frac{1}{2}P\beta,\\
\textrm{s.t.} & \ \beta^Ty=0,\\
& \ 0\leq\beta_i\leq \mu,i=1,\dots,N
\label{eq:TVSVMOCD}
\end{aligned}
\end{eqnarray}
where $Q=YGY$, $P=re^T(G+G^T)Y$ and $G=(\lambda I+rK)^{-1}K$. The above problem can be solved by gradient descent method, and the solution $\beta^*$ can be used to obtain the optimal $\alpha^*$, which is:
\begin{eqnarray}
\alpha^*=(\lambda I+rK)^{-1}(Y\beta^*+re)
\label{eq:newalpha2}
\end{eqnarray}
and
\begin{eqnarray}
f=K\alpha^*
\label{eq:newf2}
\end{eqnarray}
This leads to the proposed iterative algorithm:
\begin{eqnarray}
(f^k)^{n+1}&=&\textrm{computed by using (\ref{eq:TVSVMOCD}), (\ref{eq:newalpha2}) and (\ref{eq:newf2})}\\
(\hat{g}^k)^{n+1}&=&\argmin_{g^k}\ \gamma \|Dg^k\| + \frac{r}{2} \|g^k-(f^k+\frac{l^k}{r})\|^2_2\\
(\bar{g}^k)^{n+1}&=&\Pi_{\sum g^k=1}(\hat{g}^k)\\
(g^k)^{n+1}&=& N.\frac{(\bar{g}^k)^{n+1}}{\|(\bar{g}^k)^{n+1}\|_2}
\label{eq:}
\end{eqnarray}
\noindent
Finally, unseen data points are classified as follows:
\begin{eqnarray}
x \in C_k \textrm{ if } f^\star_k(x) = \max_j (\{f^\star_j(x)\}_{1\leq j \leq c})
\label{eq:}
\end{eqnarray}


\subsection{Cheeger-based SVM}

The Cheeger-based SVM with slack variable problem for multi-class classification is as follows: 
\begin{eqnarray}
\min_{\vec{f}=(f^1,...,f^c)\in H_K}\ \sum_{k=1}^c \frac{\sum_{i,j\in N} w_{i,j} |f^k_i-f^k_j|}{\sum_{i\in N}|f^k_i-median(f^k)|}\ s.t.\ f^k_i=y^k_i,\ \forall i\in n
\label{eq:CheegerRLSMC}
\end{eqnarray}
The following algorithm is proposed:
\begin{eqnarray}
(g^k)^{n+1} &=& (f^k)^n + c.sign((f^k)^n)\\
(e^k)^{n+1} &=& \textrm{SVM}((g^k)^{n+1})\\
(h^k)^{n+1} &=& \argmin_{h^k} \ TV(h^k) + \frac{E^n}{2c} \|h^k-(e^k)^{n+1}\|^2_2\\
(t^k)^{n+1} &=& (h^k)^{n+1} - median((h^k)^{n+1})\\
(s^k)^{n+1} &=& 
\left\{
\begin{array}{lll}
y^k(i) & \forall i\in n\\
(t^k)^{n+1}(i) & \forall i\not\in n
\end{array}
\right.
\\
(\hat{s}^k)^{n+1}&=&\Pi_{\sum s^k=1}(s^k)\\
(f^k)^{n+1} &=& N.\frac{(\hat{s}^k)^{n+1}}{\|(\hat{s}^k)^{n+1}\|_2}
\label{eq:AlgoCheegerSVMMC}
\end{eqnarray}
where SVM($\cdot$) is as (\ref{eq:CheegerSVM}).

\noindent
Finally, unseen data points are classified as follows:
\begin{eqnarray}
x \in C_k \textrm{ if } f^\star_k(x) = \max_j (\{f^\star_j(x)\}_{1\leq j \leq c})
\label{eq:}
\end{eqnarray}


\subsection{Experimental results}

\begin{table}[h!]
\vspace{0.5cm}
\centering
\begin{tabular}{|c|c|c|c|c|c|c|c|c|c|}
\hline	
\# labels per class &  1 & 5 & 10 & 50  \\
\hline	
Lap-RLS & 20.06 & 6.64 & 4.03 & 3.3 \\
\hline	
Lap-SVM & 49.95 & 14.21 & 6.27  & 2.82 \\
\hline
TV-RLS  & 2.0 & 2.06 & 1.91 & 1.98 \\
\hline
TV-SVM & {\bf 1.75} & {\bf 1.82} & 1.77 & 1.85 \\
\hline	
Cheeger-RLS  & 3.35  & 1.95 & 1.85  & 1.87 \\
\hline
Cheeger-SVM & 2.94 & 2.08 & {\bf 1.72} & {\bf 1.74 }\\
\hline	
\end{tabular}
\caption{Multi-class semi-supervised classification algorithms tested on four classes (0's, 1's, 4's and 9's) from USPS dataset. Error is averaged over 10 runs with randomly selected labels.}
\label{tab:}
\end{table}

\bibliographystyle{plain}
\bibliography{bib_sstrans}

\end{document}